\documentclass[10pt,twocolumn,letterpaper]{article}

\usepackage{iccv}
\usepackage{times}
\usepackage{epsfig}
\usepackage{graphicx}
\usepackage{amsmath}
\usepackage{amssymb}
\usepackage{pifont}
\usepackage{multirow}
\usepackage{xcolor}
\usepackage{bm}
\usepackage{graphicx}
\usepackage{booktabs}
\usepackage{color, colortbl}
\usepackage{lipsum}
\usepackage{subcaption}
\usepackage{here}

\usepackage[normalem]{ulem}
\definecolor{Gray}{gray}{0.9}
\DeclareMathOperator*{\E}{\mathbb{E}}
\newcommand{\ours}{OVANet\xspace}
\newcommand{\oursv}{OVAN\xspace}

\usepackage[pagebackref=true,breaklinks=true,letterpaper=true,colorlinks,bookmarks=false]{hyperref}

\iccvfinalcopy 

\def\iccvPaperID{5753} 

\ificcvfinal\fi

\begin{document}

\title{OVANet: One-vs-All Network for Universal Domain Adaptation}

\author{%
  Kuniaki Saito$^{1}$, \ Kate Saenko$^{1,2}$\\
  $^{1}$Boston University, $^{2}$MIT-IBM Watson AI Lab\vspace{.7em}\\
\texttt{[keisaito,saenko]@bu.edu}}

\maketitle
\ificcvfinal\thispagestyle{empty}\fi

\begin{abstract}
Universal Domain Adaptation (UNDA) aims to handle both domain-shift and category-shift between two datasets, where the main challenge is to transfer knowledge while rejecting “unknown” classes which are absent in the labeled source data but present in the unlabeled target data. Existing methods manually set a threshold to reject "unknown” samples based on validation or a pre-defined ratio of “unknown” samples, but this strategy is not practical. In this paper, we propose a method to learn the threshold using source samples and to adapt it to the target domain. Our idea is that a minimum inter-class distance in the source domain should be a good threshold to decide between “known” or “unknown” in the target. To learn the inter- and intra-class distance, we propose to train a one-vs-all classifier for each class using labeled source data. Then, we adapt the open-set classifier to the target domain by minimizing class entropy. The resulting framework is the simplest of all baselines of UNDA and is insensitive to the value of a hyper-parameter, yet outperforms baselines with a large margin. Implementation is available at \url{https://github.com/VisionLearningGroup/OVANet}.
\end{abstract}


\vspace{-3mm}
\section{Introduction}
Deep neural networks can learn highly discriminative representations for image recognition tasks~\cite{imagenet, simonyan2014very, krizhevsky2012imagenet, faster, maskrcnn} given a large amount of training data, but do not generalize well to novel domains. Collecting a large amount of annotated data in novel domains incurs a high annotation cost. To tackle this issue, domain adaptation transfers knowledge from a label-rich training domain to a label-scarce novel domain~\cite{ben2010theory}. 
Traditional unsupervised domain adaptation (UDA) assumes that the source domain and the target domain completely share the sets of categories, \ie, closed-set DA. But, this assumption does not often hold in practice. 
There are several possible situations: the target domain contains categories absent in the source (unknown categories), \ie, open-set DA (ODA)~\cite{busto2017open,saito2018open}; the source domain includes categories absent in the target (source private categories); \ie, partial DA (PDA)~\cite{cao2018partial}; a mixture of ODA and PDA, called open-partial DA (OPDA). 
Many approaches have been tailored for a specific setting, but a true difficulty is that we cannot know the category shift in advance. 

The task of universal domain adaptation (UNDA) was proposed~\cite{UDA_2019_CVPR, saito2020universal} to account for the uncertainty about the category-shift. The assumption is that the label distributions of labeled and unlabeled data can be different, but we cannot know the difference in advance. Since estimating the label distributions of unlabeled data is very hard in real applications, the setting is very practical. Although we focus on a domain shift problem in this paper, the setting also applies to a semi-supervised learning problems~\cite{grandvalet2005semi}. 

In this task, our goal is to have a model that can categorize target samples into either one of the correct known labels or the unknown label. 
The main technical difficulty is that no supervision is available to distinguish unknown samples from known ones; that is, we do not know how many of them are unknown nor the properties of the unknown instances. Obtaining this prior knowledge without manual labeling is hard in practice. 

\begin{figure}
    \centering
    \includegraphics[width=1.05\linewidth]{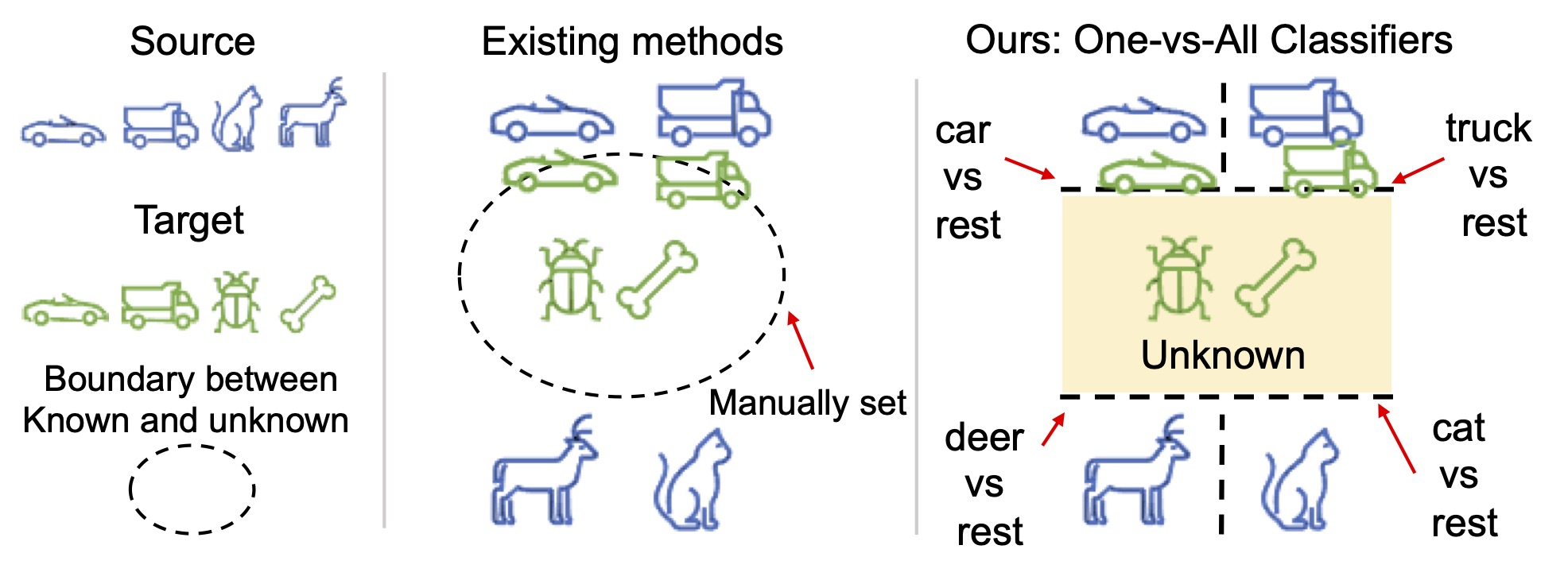}
    \vspace{-7mm}
    \caption{{\small Existing open-set or universal domain adaptation methods handle unknown samples by manually setting a threshold to reject them, either by validation or prior knowledge about the target domain. If set incorrectly, it can mistakenly reject known classes as shown here, \eg, car and truck. Instead, we propose to learn the threshold by training one-vs-all classifiers for each class.}}
    \label{fig:fig1}
    \vspace{-3mm}
\end{figure}

To allow a model to learn the concept of unknown, existing UNDA and ODA methods employ various techniques: rejecting a certain ratio of target samples~\cite{bucci2020effectiveness}, validating a threshold to decide unknown by using labeled target samples~\cite{fu2020learning}, and synthetically generating unknown instances~\cite{kundu2020universal}.
Rejecting some ratio of target samples works well if the ratio is accurate.
But, estimating the ratio is hard without having labeled target samples. Validation with labeled target samples violates the assumption of UNDA. Synthesized unknown instances define the concept of the unknown for a learned model, but tuning the generation process requires validation with labeled samples since the generated data is not necessarily similar to real unknown data. In summary, as the center of Fig. ~\ref{fig:fig1} describes, these existing methods manually define the threshold to reject unknown instances. 
To achieve a practical solution, we need an approach that does not need the ratio of unknown samples nor any validation to set the threshold. 

We cast a question to solve the problem: can we leverage the inter-class distance between source categories to learn the threshold? 
We assume that the minimum inter-class distance is a good threshold to determine whether a sample comes from the class since it defines a minimum margin from other classes. If the distance between a sample and a class is smaller than the margin, the sample should belong to the class. If the sample does not lie within the margin for any classes, it should be unknown. Fig. ~\ref{fig:fig1} illustrates the idea. Car and truck share similar features but belong to different classes. If a model knows the margin between the two classes, it can distinguish unknown classes, \eg, bug and bone, from car and truck. 

Given this insight, we explore a simple yet powerful idea: training a one-vs-all (OVA) classifier for every class in the source domain. 
We train the classifier to categorize inputs other than the corresponding class as negatives. The classifier learns a boundary between positive and negative classes, \ie, employs inter-class distance to learn the boundary. If all of the classifiers regard the input as negative, we assume the input is from unknown classes. Therefore, a model can learn the threshold to reject unknown classes by using source samples. 
In addition, we propose novel hard-negative classifier sampling, which updates open-set classifiers of a positive and a hard negative class for each source sample, to efficiently learn the minimum inter-class distance for each class. The technique makes a model scalable to a large number of classes. 
For unlabeled target samples, we propose to apply open-set entropy minimization (OEM), where the entropy of the one-vs-all classifiers is minimized. This allows a model to align unlabeled target samples to either known or unknown classes. 
Our method is significantly simpler than existing methods since it has only one unique hyper-parameter that controls the trade-off between the classification loss of source samples and OEM, yet shows great robustness to different label distributions of the target domain. 

In experiments, we extensively evaluate our method on universal domain adaptation benchmarks and vary the proportion of shared and unknown classes. This simple method outperforms various baselines that explicitly or implicitly employ the ratio of unknown samples. Moreover, the proposed way of detecting unknown classes is effective to set a threshold to reject unknown classes in semi-supervised learning. 
\begin{table}[t]
\begin{center}
\scalebox{0.85}{

\begin{tabular}{l|cccccc|c}
\toprule[1.5pt]
Method&  No. of HP & Threshold\\\hline
UAN~\cite{UDA_2019_CVPR}          &2&Validated\\
CMU~\cite{fu2020learning}          &3 &Validated\\ 
USFDA~\cite{kundu2020universal} &  3 & Synthesize unknown samples\\
ROS~\cite{bucci2020effectiveness}          &4 &Reject 50\% of target data\\
DANCE~\cite{saito2020universal}	& 3 &Decide by No. of classes \\\hline
\ours         &1 & Learned by source\\
  \bottomrule[1.5pt]
\end{tabular}}
\end{center}
\vspace{-7mm}
\caption{\small \textbf{Comparison of open-set and universal DA methods}. HP denotes hyper-parameter.
Note that USFDA~\cite{kundu2020universal} leverages synthetically generated negatives, which requires a complicated process to generate them.}
\label{tb:difference}
\end{table}
\section{Related Work}\label{sec:related}
\begin{figure*}
    \centering
    \includegraphics[width=\linewidth]{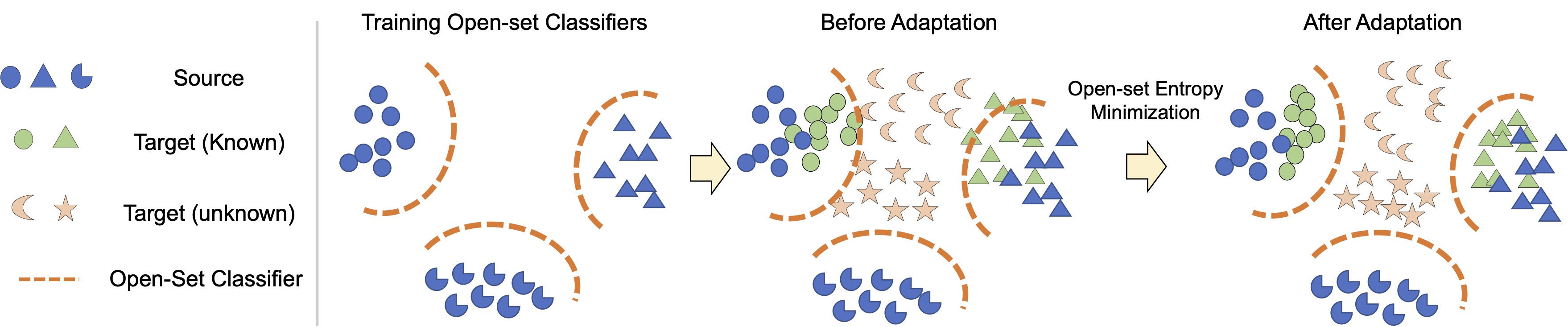}
        \vspace{-5mm}
    \caption{{\small  \textbf{Conceptual overview of \ours}. Open-set classifiers are trained on labeled source samples (leftmost), and we attempt to increase the confidence in the predictions by open-set entropy minimization on unlabeled target samples (middle and rightmost). We show the training procedure in a step-by-step manner for better understanding but employ end-to-end training in practice.}}

    \label{fig:overview}
            \vspace{-3mm}

\end{figure*}
\textbf{Domain Adaptation.}
Unsupervised domain adaptation (UDA)~\cite{saenko2010} aims to learn a good classifier for a target domain given labeled source and unlabeled target data. 
Let $L_s$ and $L_t$ denote the label space of a source and a target domain respectively. 
A closed-set domain adaptation ($L_s = L_t$) is a popular task in UDA, and distribution alignment~\cite{ganin2014unsupervised, tzeng2017adversarial,long2015learning} is one of the popular approaches. 
Open-set DA (presence of target-private classes, $|L_t - L_s| > 0, |L_t \cap L_s| = |L_s| $)~\cite{panareda2017open, saito2018open}, and partial DA (presence of source-private classes $|L_s - L_t| > 0, |L_t \cap L_s| = |L_t| $)~\cite{cao2018partial} are proposed to handle the category mismatch problem. 
Universal DA (UNDA)~\cite{UDA_2019_CVPR} is proposed to handle the mixture of these settings. Saito \etal~\cite{saito2020universal} emphasize the importance of measuring the robustness of a model to various category-shifts since we cannot know the detail of the shifts in advance. 
Prior works~\cite{UDA_2019_CVPR,fu2020learning,saito2020universal} compute a confidence score for known classes, and samples with a score lower than a threshold are regarded as unknown. Fu \etal~\cite{fu2020learning} seem to validate the threshold using labeled data, which is not a realistic solution. Bucci \etal~\cite{bucci2020effectiveness} set the mean of the confidence score as the threshold, which implicitly rejects about half of the target data as unknown.
Saito \etal~\cite{saito2020universal} set a threshold decided by the number of classes in the source, which does not always work well. We summarize how our approach, \ours, is different from existing methods in Table~\ref{tb:difference}. Our approach trains a model to learn the threshold by using source samples and attempts to adapt the threshold to a target domain. Our model is trained in an end-to-end manner, requires only one hyper-parameter, and is not sensitive to its value.

\textbf{Open-set recognition. }
Open-set recognition~\cite{bendale2016towards} handles both known and unknown samples in the test phase given known samples during training. 
Many methods focus on how to build a better measurement of anomaly~\cite{bendale2016towards} or to let a model learn features effective to distinguish known or unknown samples~\cite{yoshihashi2019classification, perera2020generative,oza2019c2ae}. Recent works show that contrastive learning is effective to learn representations suitable to detect out of distribution samples~\cite{tack2020csi, winkens2020contrastive}. 
These methods have a common issue with existing UNDA and ODA methods. The threshold to determine unknown is validated or pre-determined~\cite{geng2020recent}. These approaches provide a metric or method of training effective to calibrate the uncertainty score, but it is still necessary to determine the threshold by validation.  
By contrast, we introduce a technique that learns the threshold using labeled samples without a need for validation.

\begin{figure}
    \centering
    \includegraphics[width=\linewidth]{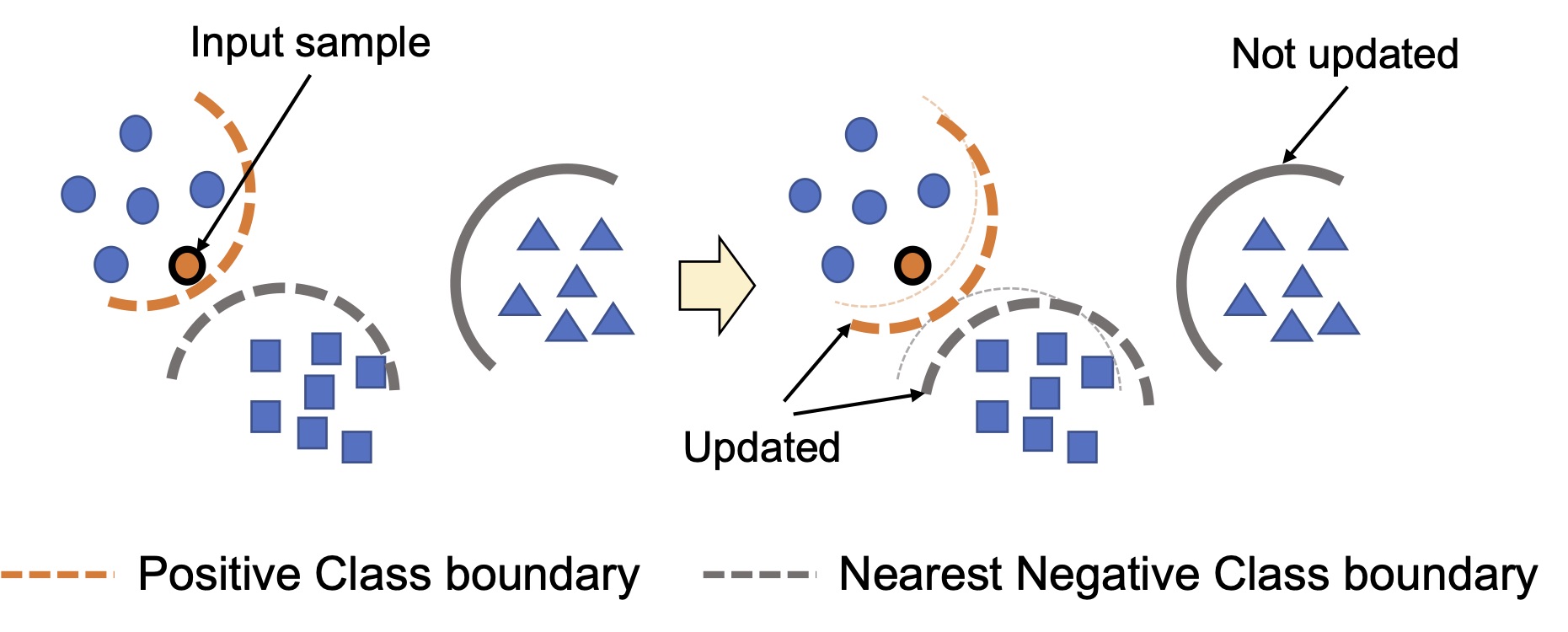}
    \vspace{-7mm}
    \caption{{\small  \textbf{Overview of the open-set classifier training.} For each sample, a positive and a nearest negative class boundary are updated. This is to let the classifier learn the distance between the positive and nearest negative class. We hypothesize that the distance should be a good threshold to reject “unknown” examples. }}
    \vspace{-3mm}

    \label{fig:oset_train}
\end{figure}
A One-vs-All (OVA) classifier is employed to apply a binary classifier to multi-class classification~\cite{galar2011overview}, aggregating the output of the binary classifier.
Padhy \etal ~\cite{padhy2020revisiting} propose to handle out-of-distribution detection by training neural OVA classifiers. A key difference from this work is that we train open-set classifiers by letting them focus on hard-negative samples. Open-set classifiers need to distinguish positive and nearest negative samples to effectively identify unknown samples. Our proposed hard-negative sampling is effective in building a threshold as shown in Sec.~\ref{sec:os_train} and Table~\ref{tb:ablation}.

\section{\ours}
Fig. ~\ref{fig:overview} introduces the conceptual overview of \ours. Our open-set classifier employs the learned distance between categories to identify unknown samples (Sec.~\ref{sec:os_train}). The learned classifiers are adapted to the target domain by open-set entropy minimization (Sec.~\ref{sec:oem}).

\textbf{Notation.} We are given a labeled source domain $\mathcal{D}_{s}=\left\{\left(\mathbf{x}_{i}^{s}, {y_{i}}^{s}\right)\right\}_{i=1}^{N_{s}}$ with ``known'' categories $L_s$ and an unlabeled target domain $\mathcal{D}_{t} = \left\{\left(\mathbf{x}_{i}^{t} \right)\right\}_{i=1}^{N_{t}}$ which contains ``known'' categories and ``unknown'' categories, where $L_s$ and $L_t$ denote the label spaces of the source and target respectively. 
Our goal is to label the target samples with either one of the $L_s$ labels or the ``unknown'' label. We train the model on $\mathcal{D}_{s} \cup \mathcal{D}_{t}$ and evaluate on $ \mathcal{D}_{t}$. To handle the open-set classification, we introduce two classifiers: (1) open-set classifiers, $O$, to detect unknown samples and (2) a closed-set classifier, $C$, to classify samples into $L_s$ labels. $C$ is trained to classify source samples with a standard classification loss while $O$ is trained with \textit{hard negative classifier sampling} explained below. In the test phase, $C$ is used to identify the nearest known class while $O$ is used to determine whether the sample is known or unknown.

\subsection{Hard Negative Classifier Sampling (HNCS)}\label{sec:os_train}
Our idea is to train a classifier to learn a boundary between in-liers and outliers for each class. Then, we train a linear classifier for each class. For each classifier, the class is trained to be positive while other classes are negative. The key is how to pick the negative samples, \ie, we propose to pick samples different from the class but similar to it (hard negative class). The overview is illustrated in Fig. ~\ref{fig:oset_train}.

Our open-set classifiers consist of $|L_s|$ sub-classifiers, i.e., $O^{j}, j \in \{1, . . . , |L_s|\}$. 
Each classifier is trained to distinguish if the sample is an in-lier for the corresponding class. The sub-classifier outputs a 2-dimensional vector, where each dimension shows the probability of a sample being an in-lier and outlier respectively. 

Padhy \etal~\cite{padhy2020revisiting} propose to train all ($|L_s|-1$) negative classifiers given a training sample, but we observe that learned classifiers are not useful when the number of classes is large, which is consistent with the observation of \cite{padhy2020revisiting}. 
The reason is that the open-set classifiers can classify many negatives too easily. Take an example of three classes: a cat, a dog, and a turtle. We assume that the features of the turtle are very different from those of the cat and the dog. 
To learn an effective boundary for the cat, a model should focus on the dog rather than the turtle since the cat should be closer to the dog in a feature space. But, if the turtle is sampled very frequently, the learned boundary can be in the middle of the cat and the turtle, accepting too many unknown samples as known. 
Therefore, each sub-classifier needs hard negative samples to learn a boundary between in-liers and outliers. 
Considering this insight, we propose to train two one-vs-all classifiers given a sample: a classifier of the positive and that of the nearest negative class (Fig.~\ref{fig:oset_train}). By picking the nearest negative class for each sample, we can let the corresponding classifier learn an effective boundary to identify unknown instances. 

We leverage a linear classifier for each sub-classifier. The open-set classifier is employed on top of an extracted feature, namely, ${\bm{z}}^{k} = {\mathbf{w}}^{k} G_{\bm{\theta}} (\mathbf{x}) \in \mathbb{R}^{2}$, where $G_{\bm{\theta}}$ and ${\mathbf{w}}^{k}$ denote a feature extractor and a weight of an open-set classifier for class $k$ respectively. Each dimension of ${\bm{z}}^{k} \in \mathbb{R}^{2}$ denotes the score for known and unknown respectively.
Let $p(\hat{y}^{k}| \mathbf{x})$ denote the output probability that the instance $\mathbf{x}$ is an in-lier for the class $k$: $p(\hat{y}^{k}| \bm{x}) = \sigma(\bm{z}^{k})_{0}$, where $\sigma$ denotes softmax activation function.

We denote $L_{ova} (\mathbf{x}^{s}, {y}^{s})$ as the open-set classification loss for a sample $(\mathbf{x}^{s}, {y}^{s})$:
\begin{equation}
\mathcal{L}_{ova} (\mathbf{x}^{s}, {y}^{s})= -\log(p(\hat{y}^{{y}^{s}} | \mathbf{x}^{s})) - \min_{j\neq {y}^{s}}\log(1-p(\hat{y}^{j} | \mathbf{x}^{s})). 
\end{equation}
This computes the loss on a positive class and the hardest negative class, as we explain above. We call the technique hard negative classifier sampling (HNCS). 
The process of computing the loss is illustrated at the top of Fig.~\ref{fig:model}.

\subsection{Open-set Entropy Minimization (OEM)}\label{sec:oem}
Given the open-set classifiers trained on the source domain, we propose to enhance the low-density separation for the unlabeled target domain. Since the target samples have different characteristics from the source, they can be classified incorrectly with respect to both closed-set and open-set categorization.
To handle the issue, we propose a novel entropy minimization method that adapts the open-set classifiers to the target domain. 

\begin{figure}
    \centering
    \includegraphics[width=\linewidth]{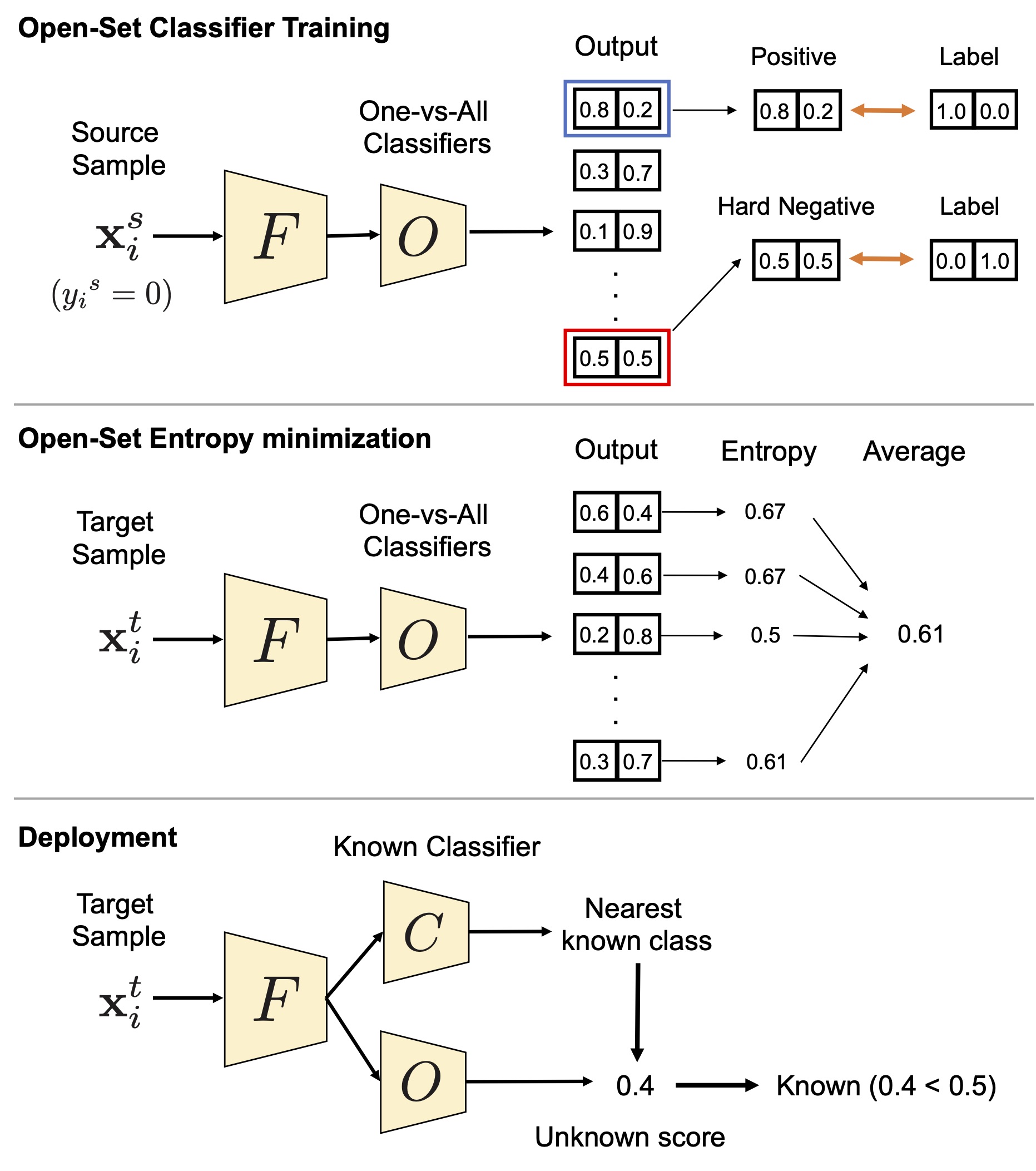}
        \vspace{-7mm}
    \caption{{\small  \textbf{Overview of training and testing.} $O$ denotes one-vs-all open-set classifiers whereas $C$ denotes a closed-set classifier. $F$ is a feature extractor.
    {\bf Top}: We propose hard-negative classifier sampling to train one-vs-all classifiers using source samples (Sec.~\ref{sec:os_train}). {\bf Middle}: We apply entropy minimization with open-set classifiers for unlabeled target samples (Sec.~\ref{sec:oem}). {\bf Bottom}: In the test phase, a nearest known class is identified by a closed-set classifier, and the corresponding open-set classifier's score is leveraged to decide known or unknown  (Sec.~\ref{sec:test}).}}

    \label{fig:model}
      \vspace{-3mm}
\end{figure}

Our idea is to increase the confidence in the prediction with regard to open-set classification, i.e., known or unknown. Specifically, we apply entropy minimization training for all open-set classifiers for each $\mathbf{x}^{t} \in \mathcal{D}_{t}$. We compute entropy of all the classifiers and take the average and train a model to minimize the entropy, as illustrated in the middle of Fig.~\ref{fig:model}.

\begin{eqnarray}
\mathcal{L}_{ent}(\mathbf{x}^{t}) =   -&\sum_{j=1}^{|L_s|}&p(\hat{y}^{j} | \mathbf{x}^{t} )\log(p(\hat{y}^{j} | \mathbf{x}^{t} )) \nonumber \\
    &+& (1 - p(\hat{y}^{j} | \mathbf{x}^{t}))\log(1-p(\hat{y}^{j} | \mathbf{x}^{t} )) \nonumber 
\end{eqnarray}
By this entropy minimization, known target samples will be aligned to source samples whereas unknown ones can be kept as unknown. One clear difference from the existing entropy minimization~\cite{grandvalet2005semi} is that we are able to keep unknown instances as unknown since the entropy minimization is performed by open-set classifiers, not by a closed-set classifier. Entropy minimization by a closed-set classifier necessarily aligns the unlabeled samples to known classes because there is no concept of an unknown class. Since our open-set classifiers have the concept of unknown, the model can increase the confidence of it. 
\begin{figure*}[t]
 \begin{subfigure}[b]{0.34\textwidth}
  \begin{center}
   \includegraphics[width=\textwidth]{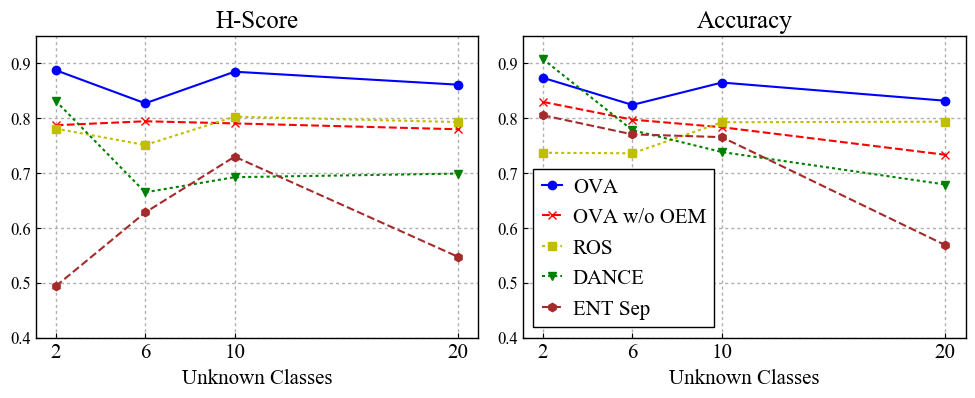}
    \caption{Amazon to DSLR}
    \label{fig:amazon2dslr}
  \end{center}
 \end{subfigure}
 ~
\begin{subfigure}[b]{0.34\textwidth}
  \begin{center}
   \includegraphics[width=\textwidth]{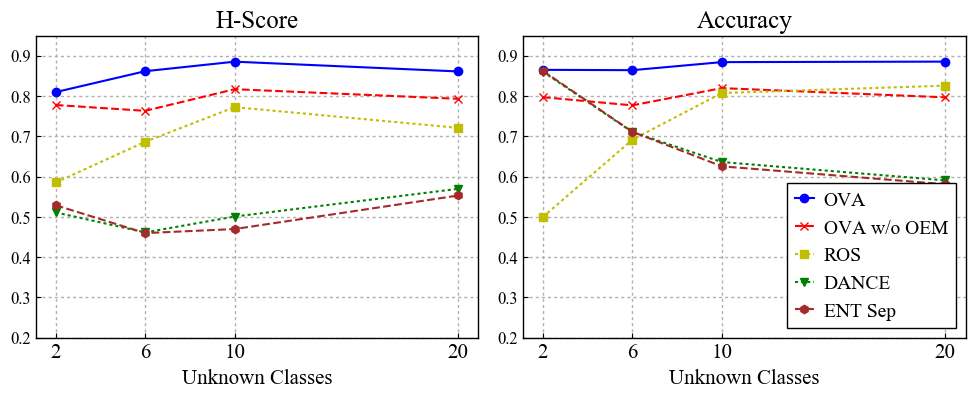}
    \caption{Webcam to Amazon}
      \label{fig:webcam2amazon}
  \end{center}
 \end{subfigure}
 ~ 
\begin{subfigure}[b]{0.34\textwidth}
  \begin{center}
   \includegraphics[width=\textwidth]{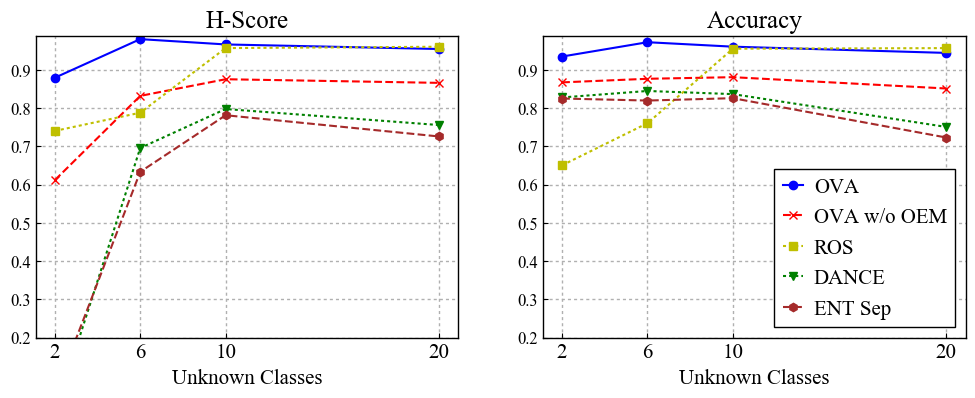}
    \caption{DSLR to Webcam}
     \label{fig:dslr2webcam}
  \end{center}
 \end{subfigure}
 \vspace{-7mm}
 
   \caption{\small \textbf{H-score and accuracy in open-set domain adaptation}. \textcolor{blue}{Blue}: \ours (Ours). We vary the number of unknown classes using Office ($|L_s \cap L_t| = 10, |L_s - L_t| = 0$). The left and right side show H-score and accuracy respectively. \ours shows stable performance across different openness while baselines can much degrade performance in some settings.}
 \label{fig:vary_openness}
   \vspace{-3mm}
\end{figure*}

\subsection{Learning}
We combine the open-set classifier and closed-set classifier to learn both open-set and closed-set categorization. For the closed-set classifier, we simply train a linear classifier on top of the feature extractor using cross-entropy loss, which we denote $L_{cls}(\bm{x}, y)$. Then, the overall training loss can be computed as follows:
\begin{eqnarray}
    \mathcal{L}_{all} = \E_{(\mathbf{x}_{i}^{s}, {y_{i}}^{s})	\sim \mathcal{D}_{s}} \mathcal{L}_{src}(\mathbf{x}_{i}^{s}, {y_{i}}^{s}) +  \lambda \E_{\mathbf{x}_{i}^{t}	\sim \mathcal{D}_{t}} \mathcal{L}_{ent}(\mathbf{x}_{i}^{t}),\\\label{eq:main}
    \mathcal{L}_{src}(\mathbf{x}_{i}^{s}, {y_{i}}^{s})=\mathcal{L}_{cls}(\mathbf{x}_{i}^{s}, {y_{i}}^{s}) + \mathcal{L}_{ova} (\mathbf{x}_{i}^{s}, {y_{i}}^{s}).
\end{eqnarray}
The parameters of $F$, $O$, and $C$ are optimized to minimize the loss.
Note that \ours has only one hyper-parameter, $\lambda$. 
This method is much simpler than existing ODA and UNDA methods~\cite{saito2020universal,fu2020learning,bucci2020effectiveness} all of which require setting the threshold manually and/or multiple training phases. 

\subsection{Inference}\label{sec:test}
In the test phase, we utilize both the trained closed-set and open-set classifier. We first get the closest known class by using the closed-set classifier and take the corresponding score of the open-set classifier. The process is illustrated at the bottom of Fig.~\ref{fig:model}.

\section{Experiments}
\vspace{-1mm}
We evaluate our method in UNDA settings along with ablation studies.
To evaluate the robustness to the change of the number of unknown target samples, we vary the number and compare it with other baselines. 
\subsection{Setup}
\textbf{Datasets.}\quad
We utilize popular datasets in DA: Office~\cite{saenko2010}, OfficeHome~\cite{venkateswara2017Deep}, VisDA~\cite{peng2017visda}, and DomainNet~\cite{peng2018moment}. Unless otherwise noted, we follow existing protocols~\cite{fu2020learning, UDA_2019_CVPR, saito2018open} to split the datasets into source-private ($|L_s - L_t|$), target-private ($|L_t - L_s|$) and shared categories ($|L_s \cap L_t|$). 
For fairness, Saito \etal~\cite{saito2020universal} propose to evaluate universal DA methods on the various number of unknown and known classes, which can reveal methods tailored for a specific setting. We follow their policy and provide experimental results varying the number of unknown and known classes. Since many existing methods are optimized to handle a specific benchmark, we aim to fairly evaluate methods with respect to the sensitivity to diverse settings. To briefly describe the method for splitting categories in each experiment, each table shows $|L_s \cap L_t| / |L_s - L_t| / |L_t - L_s|$, \ie, (shared, source private, and target private classes).  See our supplementary material for more details.

\textbf{Evaluation Metric.}\quad
Considering the trade-off between the accuracy of known and unknown classes is important in evaluating universal or open-set DA methods. 
To this end, we evaluate methods using H-score~\cite{bucci2020effectiveness}. 
H-score is the harmonic mean of the accuracy on common classes ($acc_c$) and accuracy on the “unknown” classes $acc_t$ as: 
\begin{equation}
    H_{score} = \frac{2 acc_c \cdot acc_t}{acc_c + acc_t.}
\end{equation}
The evaluation metric is high only when both the $acc_c$ and $acc_t$ are high. So, H-score can measure both accuracies of UNDA methods well. Unless otherwise noted, we show H-score in tables and graphs. 
The drawback of this metric is that the importance of recognizing known and unknown classes is always equal. If the number of unknown instances is small, the metric puts too much weight on the unknown class. Thus, when the number is small, we also report the instance-wise accuracy over all samples. 

\begin{table}[t]
 \addtolength{\tabcolsep}{-3pt}
\begin{center}
\scalebox{0.9}{

\begin{tabular}{l|cccccc|c}
\toprule[1.5pt]
\multirow{2}{*}{Method}& \multicolumn{6}{c|}{Office (10 / 10 / 11)}&\multirow{2}{*}{Avg}\\
     & A2D  & A2W  & D2A  & D2W  & W2D  & W2A  &   \\\hline
UAN~\cite{UDA_2019_CVPR}          & 59.7 & 58.6 & 60.1 & 70.6 & 71.4 & 60.3 & 63.5 \\
CMU~\cite{fu2020learning}          & 68.1 & 67.3 & 71.4 & 79.3 & 80.4 & 72.2 & 73.1 \\
DANCE~\cite{saito2020universal}	&78.6&71.5&79.9&91.4&87.9&72.2&80.3 \\
DCC~\cite{lidomain} &\bf{88.5}&78.5&70.2&79.3&88.6&75.9&80.2\\
ROS~\cite{bucci2020effectiveness}          & 71.4 & 71.3 & 81.0 & 94.6 & \bf{95.3} & 79.2 & 82.1 \\
USFDA~\cite{kundu2020universal} &85.5&\bf{79.8}&\bf{83.2}&90.6&88.7&81.2&84.8\\\hline
\rowcolor{Gray}
\ours w/o OEM &69.6&63.1&79.9&85.9&88.7&80.6&77.9     \\
\rowcolor{Gray}
\ours         & 85.8 & 79.4 & 80.1 & \bf{95.4} & 94.3 & \bf{84.0} & \bf{86.5}\\
  \bottomrule[1.5pt]

\end{tabular}}
\end{center}
\vspace{-7mm}
\caption{\small \textbf{Open-partial domain adaptation using Office (H-score).}}
\label{tb:office_opda}

\end{table}

\begin{figure*}[h!]
 \begin{subfigure}[b]{0.19\textwidth}
  \begin{center}
   \includegraphics[width=\textwidth]{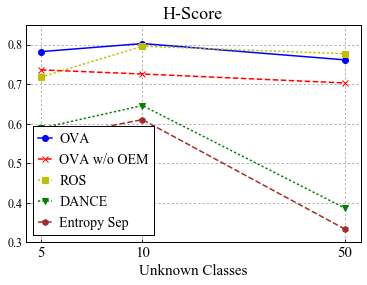}
    \caption{Real to Art}
    \label{fig:real2art}
  \end{center}
 \end{subfigure}
 ~
\begin{subfigure}[b]{0.19\textwidth}
  \begin{center}
   \includegraphics[width=\textwidth]{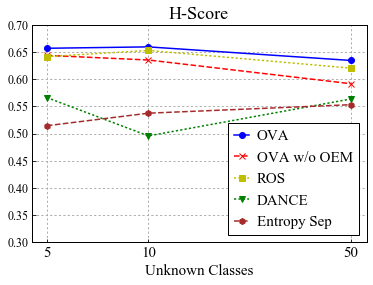}
    \caption{Real to Clipart}
      \label{fig:real2clipart}
  \end{center}
 \end{subfigure}
 ~ 
\begin{subfigure}[b]{0.19\textwidth}
  \begin{center}
   \includegraphics[width=\textwidth]{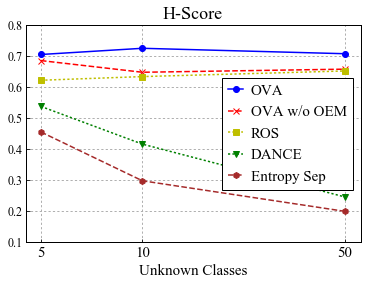}
    \caption{Clipart to Art}
     \label{fig:clipart2art}
  \end{center}
 \end{subfigure}
 ~
 \begin{subfigure}[b]{0.19\textwidth}
  \begin{center}
   \includegraphics[width=\textwidth]{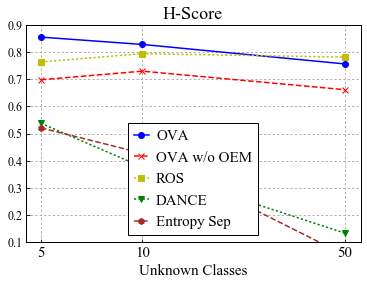}
    \caption{Art to Product}
      \label{fig:art2product}
  \end{center}
 \end{subfigure}
 ~ 
 \begin{subfigure}[b]{0.19\textwidth}
  \begin{center}
   \includegraphics[width=\textwidth]{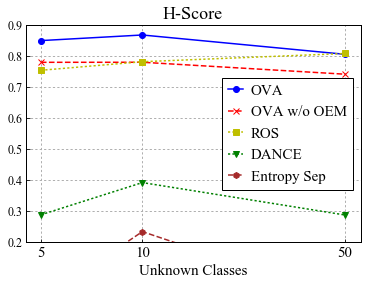}
    \caption{Product to Real}
      \label{fig:product2real}
  \end{center}
 \end{subfigure}
 \vspace{-7mm}
 
   \caption{\small \textbf{H-score of open-partial domain adaptation}. \textcolor{blue}{Blue}: \ours (Ours). We vary the number of unknown classes in OfficeHome ($|L_s \cap L_t| = 10, |L_s - L_t| = 5$).  \ours shows stable performance across different openness while baselines can much degrade performance in some settings.}
 \label{fig:officehome}
   \vspace{-2mm}
\end{figure*}

\begin{table*}[t]

\begin{center}
\scalebox{0.9}{

\begin{tabular}{l|cccccc|c|c|c}
\toprule[1.5pt]
\multirow{2}{*}{Method}& \multicolumn{6}{c|}{DomainNet (150 / 50 / 145)}&& VisDA & OfficeHome \\
    & P2R  & R2P  & P2S  & S2P  & R2S  & S2R  & Avg  & (6 / 3 / 3) & (15 / 5 / 50)\\\hline
DANCE~\cite{saito2020universal} &21.0&47.3&37.0&27.7&\bf{46.7}&21.0&33.5  & 4.4&49.2 \\
UAN~\cite{UDA_2019_CVPR}  &41.9& 43.6&39.1&38.9& 38.7& 43.7 &41.0& 30.5 &56.6\\
CMU~\cite{fu2020learning}   & 50.8 & \bf{52.2} &45.1 & 44.8 & 45.6 & 51.0 & 48.3  &34.6&61.6\\
DCC~\cite{lidomain}&\bf{56.9} & 50.3&43.7&44.9&43.3& 56.2& 49.2 & 43.0&70.2\\
\rowcolor{Gray}

\ours  & 56.0 & 51.7 & \bf{47.1} & \bf{47.4} & 44.9 & \bf{57.2} &  \bf{50.7}&\bf{53.1} &\bf{71.8}\\
  \bottomrule[1.5pt]
\end{tabular}
}
\vspace{-3mm}
\caption{\small \textbf{H-score of open-partial DA using DomainNet, VisDA and OfficeHome}. Note that CMU~\cite{fu2020learning} and DCC~\cite{lidomain} use different hyper-parameters for different datasets while \ours uses the same hyper-parameter throughout all settings. }

\label{tb:domainnet_opda}
\end{center}
\vspace{-7mm}
\end{table*}

\textbf{Implementation.} 
Following previous works~\cite{saito2020universal, UDA_2019_CVPR}, we employ ResNet50~\cite{he2016deep} pre-trained on ImageNet~\cite{imagenet} as our backbone network. Evaluation using VGGNet~\cite{simonyan2014very} is also performed in analysis. We replace the last linear classification layer with a new linear layer. 
We follow ~\cite{saito2020universal} and train our models with inverse learning rate decay scheduling. Note that we set $\lambda = 0.1$ across all settings. The value is determined by the result of open-set DA using Office (Amazon to DSLR) following  DANCE~\cite{saito2020universal}.

\textbf{Baselines.}\quad
We aim to compare methods of universal domain adaptation, which are able to reject unknown samples, such as DANCE~\cite{saito2020universal}, UAN~\cite{UDA_2019_CVPR}, ROS~\cite{bucci2020effectiveness} and CMU~\cite{fu2020learning}. To see the difference from using closed-set classifier's entropy as a threshold, we employ Entropy Separation (Ent Sep) ~\cite{saito2020universal} in several experiments. Note that these baselines are unfair in that they utilize validated thresholds or heuristically decided thresholds. 
We decide not to include the results of standard domain alignment baselines such as DANN~\cite{ganin2014unsupervised}, CDAN~\cite{long2017conditional} since existing works have already shown that these methods significantly worsen the performance in rejecting unknown samples.  
Since CMU~\cite{fu2020learning} does not publish complete code to reproduce the results, we rely on their reported results to compare with the method.

\begin{table*}[ht]
 \addtolength{\tabcolsep}{-3pt}
\begin{center}
\scalebox{0.9}{
\begin{tabular}{cc|cccc|cccc|cccc}
\toprule[1.5pt]
\multicolumn{2}{c|}{Ablation}&\multicolumn{4}{c|}{Office (W2A) (10 / 0 / 11)} &\multicolumn{4}{c|}{OfficeHome (R2A) (20 / 0 / 45)}& \multicolumn{4}{c}{DomainNet (R2P) (150 / 50 / 145)}\\
              HNCS&OEM   & H-score & Acc close & UNK   & AUROC & H-score & Acc close & UNK &  AUROC   & H-score & Acc close & UNK   & AUROC \\\hline
&\checkmark	&84.7&94.4	&82.7&92.4&63.5&80.7&58.0&76.3&	18.4&	52.2&	11.3 & 66.1\\

\checkmark&	&79.1	&93.3	&83.1 &89.4&64.9&79.9&62.5&75.6&49.7&50.7&57.8&67.2 \\
\checkmark&\checkmark&\bf{87.4}&\bf{94.7}&\bf{90.3}&\bf{94.7}&\bf{67.5}&\bf{81.3}&\bf{68.4}&\bf{77.6}&\bf{51.9}&\bf{53.5}&\bf{74.4} &\bf{67.6} \\
  \bottomrule[1.5pt]

\end{tabular}
}
\end{center}
\vspace{-7mm}
\caption{\small \textbf{Ablation study.} We ablate open-set entropy minimization (OEM) and/or hard negative classifier sampling (HNCS). Note that both techniques are necessary to achieve good performance (FULL vs each ablation).}
\label{tb:ablation}
\vspace{-3mm}

\end{table*}

\begin{figure*}[t]
 \begin{subfigure}[b]{0.34\textwidth}
  \begin{center}
   \includegraphics[width=\textwidth]{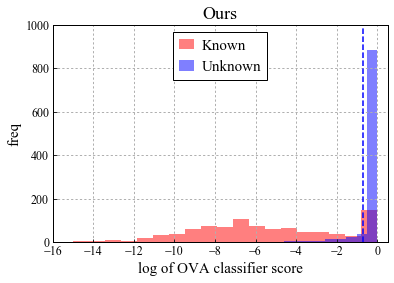}
    \caption{\ours}
    \label{fig:hist_ours}
  \end{center}
 \end{subfigure}
 ~
\begin{subfigure}[b]{0.34\textwidth}
  \begin{center}
   \includegraphics[width=\textwidth]{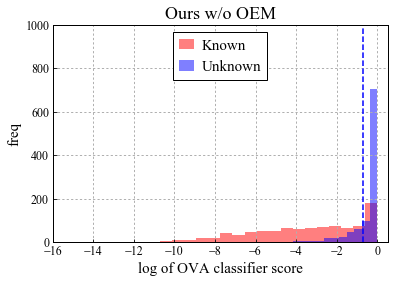}
    \caption{\ours w/o OEM}
      \label{fig:hist_ours_noadapt}
  \end{center}
 \end{subfigure}
 ~ 
\begin{subfigure}[b]{0.34\textwidth}
  \begin{center}
   \includegraphics[width=\textwidth]{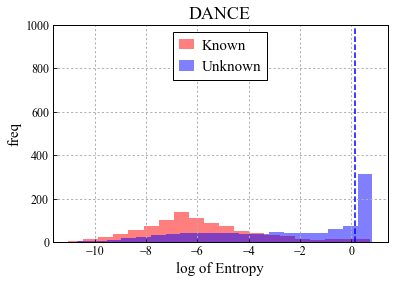}
    \caption{DANCE}
     \label{fig:hist_dance}
  \end{center}
 \end{subfigure}
 \vspace{-7mm}
 
   \caption{\small \textbf{Histogram of anomaly score in open-set DA} (Webcam to Amazon, $|L_s \cap L_t| = 10, |L_s - L_t| = 0, |L_t - L_s| = 11$). \textcolor{red}{Red}: Known samples. \textcolor{blue}{Blue}: Unknown samples. The output of open-set classifiers is used as anomaly score in \ours, and the entropy of a closed classifier is used for DANCE. Note that the threshold between known and unknown is a blue dotted line in each graph. \ours separates known and unknown samples well whereas DANCE fails.}
   \label{fig:histogram}
   \vspace{-5mm}
\end{figure*}

\textbf{Overview of Results.} \quad
In summary, our method is superior or comparable to baseline methods across all diverse settings without optimizing the hyper-parameter for each setting. The fact is verified by diverse settings using 4 benchmark datasets.  

\textbf{Office.} \quad
Fig. ~\ref{fig:vary_openness} shows the results of varying the number of unknown classes in the ODA setting on Office. The left and the right of each graph show H-score and instance-wise accuracy respectively. The x-axis denotes the number of unknown classes.
The number varies from 2 to 20 while that of known classes is fixed to 10. The performance of \ours is always better than or comparable to baselines. Even without OEM, our proposed model consistently performs well. Comparison between \ours and \ours w/o OEM demonstrates the effectiveness of OEM. 
Since ROS~\cite{bucci2020effectiveness} sets a threshold to reject about half of the target samples as unknown, it performs poorly when the unknown samples are rare.
DANCE~\cite{saito2020universal} is also sensitive to the number of unknown classes.
Table~\ref{tb:office_opda} shows the result of the OPDA setting on Office, where \ours outperforms baselines on average. 

\textbf{OfficeHome, VisDA and DomainNet.}  \quad
Results of varying the openness in the OPDA on OfficeHome are described in Fig.~\ref{fig:officehome}. We pick 5 adaptation scenarios to cover various domains and vary the number of unknown classes. The trend is similar to experiments on Office. \ours consistently performs better than baselines.
Results of the OPDA on OfficeHome, VisDA and DomainNet are summarized in Table~\ref{tb:domainnet_opda}, where we follow CMU~\cite{fu2020learning} to split the categories. For OfficeHome, the mean of 12 adaptation scenarios is presented.
In VisDA and DomainNet, the number of samples and/or that of classes are very different from those of Office and OfficeHome. 
\ours outperforms existing methods with a large margin, more than 10 points in VisDA and OfficeHome. 
Note that CMU~\cite{fu2020learning} selects the optimal threshold hyper-parameters for each dataset, while \ours achieves the best H-score without tuning a hyper-parameter. 
From these results, we observe that \ours works well across diverse settings. 

\subsection{Analysis in Domain Adaptation}
\textbf{How effective are OEM and HNCS?}
Table ~\ref{tb:ablation} shows ablation study in ODA on Office (Webcam to Amazon), ODA on OfficeHome (Real to Art), and OPDA on DomainNet (Real to Painting). \textit{Acc close} measures accuracy at recognizing known samples without rejection, i.e, this metric evaluates the ability of closed-set recognition. \textit{UNK} is the accuracy of rejecting unknown samples. \textit{AUROC} measures how well known and unknown samples are separated given the open-set classifier's output.
We have two observations: applying open-set entropy minimization (OEM) is effective to improve both closed-set accuracy and rejecting unknown examples (w/o OEM vs FULL), and hard-negative classifier selection is an appropriate way of training the classifier (w/o HNCS vs FULL). The effectiveness of hard-negative sampling is more evident in DomainNet because the dataset has much more known classes. When the number of known classes is large, there are many useless negative classes to train our open-set classifier. Then, without the sampling, the model cannot learn an effective decision boundary between known and unknown, resulting in a significant degradation in the unknown sample recognition.

\textbf{Unknown and known samples are well separated by OVA classifiers.}
Fig.~\ref{fig:histogram} shows histograms of anomaly score, where the x-axis is the anomaly score, and the y-axis shows the frequency of the corresponding range of anomaly score. The blue dotted line denotes the threshold between known and unknown.
(b) indicates that the learned threshold works well to separate known and unknown samples. Applying OEM further enhances the separation ((a) vs (b)). (c) DANCE utilizes a manually set threshold and fails in rejecting many unknown samples.
\begin{figure}[t]
\begin{subfigure}[b]{0.2\textwidth}
  \begin{center}
   \includegraphics[width=\textwidth]{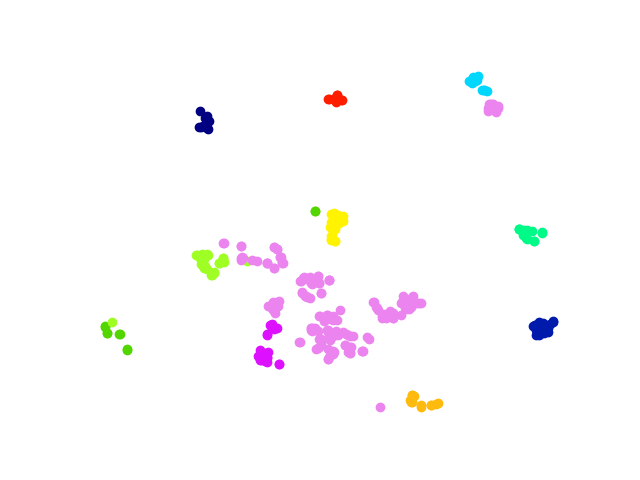}
    \caption{Ground truth labels}
      \label{fig:tsne_target}
  \end{center}
 \end{subfigure}
 ~ 
\begin{subfigure}[b]{0.2\textwidth}
  \begin{center}
   \includegraphics[width=\textwidth]{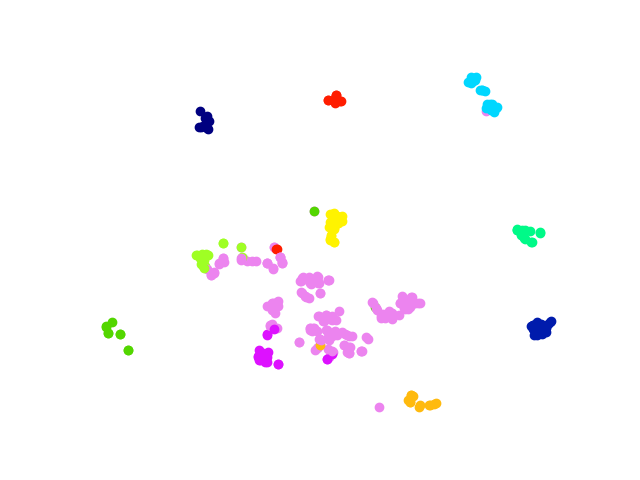}
    \caption{Predictions by \ours}
     \label{fig:tsne_decision}
  \end{center}
 \end{subfigure}
 \vspace{-3mm}
   \caption{\small \textbf{Feature visualization} with t-SNE~\cite{maaten2008visualizing}. Comparison between (a) true labels and (b) decision by \ours (ODA, Amazon to DSLR). Different colors indicate different classes. Plots with pink are unknown samples, others are known samples. }
   \label{fig:tsne}
 \vspace{-2mm}

\end{figure}

\begin{figure}[ht]
	\centering
	\begin{subfigure}{0.235\textwidth}
	\includegraphics[width=\linewidth]{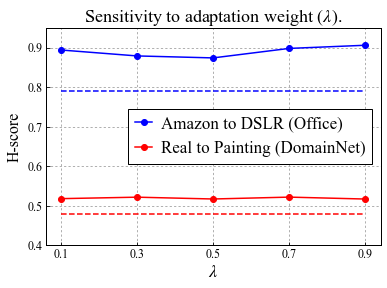}
	\caption{Sensitivity to $\lambda$}
	  \label{fig:lambda_sensitivity}
	\end{subfigure}
	\begin{subfigure}{0.235\textwidth}
	\includegraphics[width=\linewidth]{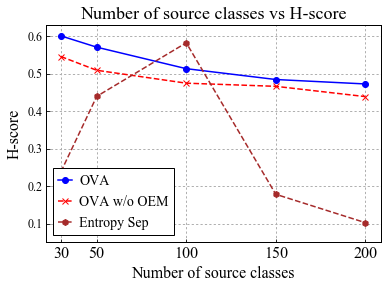}
	\caption{Sensitivity to $|L_s|$}
	 \label{fig:class_increase}
	\end{subfigure}
	\vspace{-4.0mm}
	\caption{\small (a) Sensitivity to a hyper-parameter $\lambda$ in Eq. \ref{eq:main}. Dotted lines are results w/o OEM. (b) Sensitivity to the number of source classes, where $|L_s \cap L_t| = 30, |L_t - L_s| = 45$ in DomainNet. }
	\label{fg:sensitivity}
\end{figure}

\textbf{Feature Visualization.}
Fig.~\ref{fig:tsne} visualizes learned features with corresponding ground-truth labels (a) and predicted labels (b). Although not all unknown samples are clearly separated from known ones, most are correctly classified. 

\textbf{OEM is not sensitive to a hyper-parameter.}
Fig.~\ref{fig:lambda_sensitivity} shows the sensitivity to the hyper-parameter $\lambda$ in ODA on Office and OPDA on DomainNet. $\lambda$ is the only hyper-parameter specific to \ours. \ours shows stable performance across different values of $\lambda$.

\textbf{Varying known classes.}
The performance across different numbers of source-private classes($|L_s|$) is visualized in Fig.~\ref{fig:class_increase}, 
As there are more out-lier source classes, correctly classifying target classes can become harder. While Entropy Separation~\cite{saito2020universal} is very sensitive to the number, \ours shows stable performance. 

\begin{figure}[t]
	\centering
	\begin{subfigure}{0.235\textwidth}
	\includegraphics[width=\linewidth]{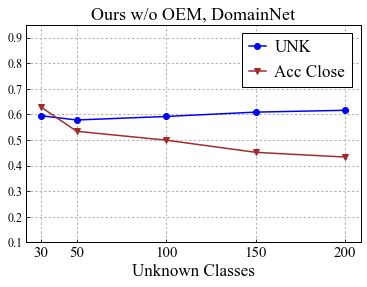}
	\caption{DomainNet}
	\end{subfigure}
	\begin{subfigure}{0.235\textwidth}
	\includegraphics[width=\linewidth]{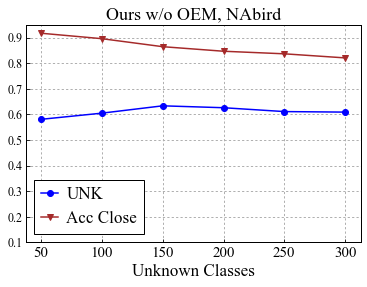}
	\caption{NAbird}
	\end{subfigure}
	\vspace{-4.0mm}
	\caption{\small \textbf{Increasing number of known classes} with \ours w/o OEM. \textcolor{blue}{Blue}: Accuracy for known samples. \textcolor{brown}{Brown}: Accuracy for unknown samples. As a model sees a larger number of known classes, the model rejects more unknown samples.}
	\label{fg:class_vs_unk}
	\vspace{-3mm}
\end{figure}

\textbf{More known classes lead to a better boundary to reject unknown.}
Fig.~\ref{fg:class_vs_unk} focuses on the results of \ours w/o OEM in increasing the number of known classes while fixing the number of unknown classes. See appendix for the detail of used datasets.
Although the number gets larger, there is more probability of unknown samples being categorized into known classes. However, the accuracy at rejecting unknown classes improves or does not change with the increase of the number. This indicates that the model is more likely to correctly reject unknown samples given more known classes. This observation is consistent with the design of one-vs-all classifiers. It is necessary for the classifiers to see hard negative samples to build a good boundary. When more classes are available, there will be more hard negatives. 

\textbf{OVANet is effective for different networks.}
Table~\ref{tb:vgg_obda} shows results of ODA on Office using VGGNet as a backbone. Note that ROS~\cite{bucci2020effectiveness} changes hyper-parameters to train their model while we do not change the hyper-parameter of \ours. The idea of employing one-vs-all classifiers is effective for different networks. 

\textbf{OVANet needs both open-set and closed-set classifiers.}
OVANet employs both open-set and closed-set classifiers, but an alternative way of training is to employ only an open-set classifier. 
However, we see a large degrade in performance if we do not use a closed-set classifier in training. This is because the open-set classifier is not trained to distinguish between known classes. The training objective considers only distinguish one known class from other classes, which is not enough to give discriminative features to classify known samples. 

\begin{table}[t]
 \addtolength{\tabcolsep}{-3pt}
\begin{center}
\scalebox{0.9}{

\begin{tabular}{l|cccccc|c}
\toprule[1.5pt]
\multirow{2}{*}{Method}& \multicolumn{6}{c|}{Office (10 / 0 / 11)}&\multirow{2}{*}{Avg}\\
     & A2D  & A2W  & D2A  & D2W  & W2D  & W2A  &   \\\hline
OSBP~\cite{saito2018open} & 81.0 & 77.5 & 78.2 & \bf{95.0} & 91.0 & 72.9 & 82.6 \\
ROS~\cite{bucci2020effectiveness}  & 79.0 & 81.0 & 78.1 & 94.4 & \bf{99.7} & 74.1 & 84.4 \\
\rowcolor{Gray}
\ours & \bf{89.5} & \bf{84.9} & \bf{89.7} & 93.7 & 85.8 & \bf{88.5} & \bf{88.7} \\
  \bottomrule[1.5pt]

\end{tabular}}
\end{center}
\vspace{-7mm}
\caption{\small Results of using \textbf{VGGNet}. Open-set DA setting is used.}
\label{tb:vgg_obda}
\vspace{-3mm}

\end{table}
 
\subsection{Semi-supervised Learning}
 
So far, we have assessed \ours in the domain adaptation scenario. However, the importance of detecting an unknown class is not limited to domain adaptation. When deploying a recognition model in a real application, a model can encounter samples from classes unseen during training. In semi-supervised learning, a model can encounter unlabeled unknown data during training. Models able to detect unknown samples are desirable. In this section, we provide the analysis of \ours for the semi-supervised setting. 

\textbf{\ours is effective in a semi-supervised setting.}
One interesting application of open-set classification and semi-supervised learning is the classification of animals. The unlabeled images of wild animals can be collected by a monitoring camera or crawling the web. Annotating the images may need expert knowledge. Therefore, the number of annotated images can be limited, and the obtained known label-space may not cover all categories of the dataset. Then, we utilize NAbird~\cite{van2015building}, a large scale bird image dataset, to evaluate our method, where 300 categories are known and 255 are unknown. Half of the samples are labeled for each known class, and the rest are treated as unlabeled data. All samples of unknown classes are used as unlabeled data. 
\ours outperforms the baselines in all metrics (Table~\ref{tb:nabird}). 
This result validates the benefit of \ours without domain-shift between labeled and unlabeled data.
\begin{table}[t]
\begin{center}
\scalebox{0.9}{

\begin{tabular}{l|cccc}
\toprule[1.5pt]
\multirow{2}{*}{Method}& \multicolumn{4}{c}{NAbird (300 / 0 / 255)}\\
& H-score& UNK & Acc & Acc close\\\hline
Ent Sep~\cite{saito2020universal} &3.8&1.9&32.2 & 81.9\\ 
DANCE~\cite{saito2020universal} &8.8&4.6&34.0 &82.1\\ 
\rowcolor{Gray}
\ours&\bf{67.6}&\bf{63.1}&\bf{67.4} & \bf{83.0}\\
  \bottomrule[1.5pt]

\end{tabular}}
\end{center}
\vspace{-7mm}
\caption{\small \textbf{Results of open-set semi-supervised learning} using NAbird~\cite{van2015building}. \textit{Acc} is the instance-wise accuracy for all samples.}
\label{tb:nabird}
  \vspace{-2mm}
\end{table}


 \section{Conclusion}
 In this paper, we present a new technique, \ours, which trains a One-vs-All classifier for each class and decides known or unknown by using the output. Our proposed framework is the simplest of all UNDA methods, yet shows strong performance across diverse settings. The extensive evaluation shows \ours's applicability to semi-supervised settings. 
\section{Acknowledgment}
This work was supported by Honda, DARPA LwLL and NSF
Award No. 1535797.

\clearpage
\renewcommand{\thefigure}{\Alph{figure}}
 \renewcommand{\thetable}{\Alph{table}}
 \def\thesection{\Alph{section}}
\setcounter{section}{0}
\setcounter{figure}{0}
\setcounter{table}{0}
\begin{table*}[h!]
\begin{center}
\scalebox{1.0}{

\begin{tabular}{l|cccccccccccc|c}
\toprule[1.5pt]
\multirow{2}{*}{Method}& \multicolumn{12}{c|}{OfficeHome (10 / 5 / 50)}&\multirow{2}{*}{Avg}\\
     & A2C&A2P&A2R&C2A&C2P&C2R&P2A&P2C&P2R&R2A&R2C&R2P   &   \\\hline
     OSBP~\cite{saito2018open} & 39.6 & 45.1 & 46.2 & 45.7 & 45.2 & 46.8 & 45.3 & 40.5 & 45.8 & 45.1 & 41.6 & 46.9 & 44.5 \\
UAN~\cite{UDA_2019_CVPR}          & 51.6 & 51.7 & 54.3 & 61.7 & 57.6 & 61.9 & 50.4 & 47.6 & 61.5 & 62.9 & 52.6 & 65.2 & 56.6\\
CMU~\cite{fu2020learning}          & 56.0&56.9&59.1&66.9& 64.2&67.8& 54.7& 51.0& 66.3& 68.2&57.8 &69.7&61.6\\
\rowcolor{Gray}
\ours         & \bf{62.8} & \bf{75.6} & \bf{78.6} & \bf{70.7} &\bf{68.8} & \bf{75.0} & \bf{71.3} & \bf{58.6} & \bf{80.5} & \bf{76.1} &\bf{64.1} & \bf{78.9} & \bf{71.8} \\

  \bottomrule[1.5pt]

\end{tabular}}
\end{center}
\vspace{-7mm}
\caption{\small \textbf{Open-partial domain adaptation on OfficeHome (H-score)}.}
\label{tb:officehome_opda}

\end{table*}
\begin{table*}[h!]
\begin{center}

\begin{tabular}{ccc|ccc|ccc}
\toprule[1.5pt]

     & A2D     &      &      & A2W     &      &      & D2A     &      \\\hline
All  & H-Score & UNK  & All  & H-Score & UNK  & All  & H-Score & UNK  \\\hline
88.3 & 90.5    & 88.0 & 87.4 & 88.3    & 86.6 & 87   & 86.7    & 95.4 \\\hline
\\\hline
     & W2A     &      &      & W2D     &      &      & D2W     &      \\\hline
All  & H-Score & UNK  & All  & H-Score & UNK  & All  & H-Score & UNK  \\\hline
88.5 & 88.3    & 91.5 & 98.3 & 98.4    & 96.9 & 97.5 & 98.2    & 98.1 \\
  \bottomrule[1.5pt]

\end{tabular}
\end{center}
\vspace{-7mm}
\caption{\small \textbf{Open-set domain adaptation on Office (10 / 0 / 11).}. ResNet50 is used as a backbone.}
\label{tb:op_office_detail}
\end{table*}
\begin{table*}[h!]


\begin{center}
\scalebox{1.0}{

\begin{tabular}{c|ccc|cccc|c}
\toprule[1.5pt]
\multirow{2}{*}{Confidence Score}& \multicolumn{3}{c|}{Office (10 / 0 / 11)} &  \multicolumn{4}{c|}{OfficeHome (10 / 5 / 50)} & DNet R2P\\
        & A2D  & D2W  & W2A  & R2A  & R2C  & A2P  & P2R  & (150 / 50 / 145)      \\\hline
Softmax & 87.6 & 97.6&85.9  &80.9&67.8&81.3&88.2&65.5\\
Entropy & 88.3 & 97.1 & 86.2 & 82.1 & 68.5 & 82.4 & 88.7 & 66.8 \\
\rowcolor{Gray}
One-vs-All Classifier  & 90.1 & 97.3 & 87.3 & 81.4 & 67.5 & 82.9 & 88.9 & 67.3 \\
  \bottomrule[1.5pt]
\end{tabular}
}
\vspace{-3mm}
\caption{\small \textbf{AUROC score.} The score measures how well known and unknown samples are separated.}
\label{tb:auroc}
\end{center}
\vspace{-7mm}
\end{table*}

In this appendix, we provide details of experiments and additional results. 
\section{Experimental Details}
\textbf{Implementation.}
The batch-size of source and target is 36 for all experiments in UNDA. 
The initial learning rate of networks is set as 0.01 for new layers and 0.001 for backbone layers in experiments on ResNet50. The learning rate is decayed with inverse learning rate decay scheduling. We follow ~\cite{bucci2020effectiveness} for setting the learning rate of VGGNet. We use Pytorch~\cite{pytorch} to implement our method. 

\textbf{Category Selection.}
After deciding shared categories and source-private categories following default benchmark settings~\cite{fu2020learning,bucci2020effectiveness}, we pick the  unknown classes in alphabetical order in Fig. 5 and 6. In Table 2, 3, and 5, we follow default benchmark settings~\cite{fu2020learning,bucci2020effectiveness}. 
In Table 6 and Fig. 10(b), which use NAbird, we set the first 300 classes as known and the rest as unknown.  
In Fig. 9(b) and Fig. 10(a), known and unknown classes of DomainNet are also picked in alphabetical order. 
For CIFAR10 and CIFAR100 (Table 7), we pick the first $N$ categories as known and rest as unknown, where $N$ denotes the number of known classes. We decide the order of categories of provided label-indexes. 

\textbf{Semi-superivised Learning.}
The implementation of the experiment is the same as UNDA, where we use ResNet50 as a backbone.
We simply replace the source with labeled data and the target with unlabeled data. 

\textbf{Open-Set Recognition.}
In this experiment, models are trained from scratch, where we employ WideResNet~\cite{zagoruyko2016wide}. We follow the implementation of Fixmatch~\cite{sohn2020fixmatch} to train the models.
Training is performed similarly to the experiments of \oursv w/o OEM except that we add contrastive loss. 

Then, the overall training loss can be computed as follows:
\begin{eqnarray}
    \mathcal{L}_{all} = \E_{(\mathbf{x}_{i}, {y_{i}})	\sim \mathcal{D}} \mathcal{L}_{src}(\mathbf{x}_{i}, {y_{i}}) +  \lambda \E_{\mathbf{x}_{i}\sim \mathcal{D}} \mathcal{L}_{simc}(\mathbf{x}_{i}),\\
    \mathcal{L}_{src}(\mathbf{x}_{i}^{s}, {y_{i}}^{s})=\mathcal{L}_{cls}(\mathbf{x}_{i}, {y_{i}}) + \mathcal{L}_{ova} (\mathbf{x}_{i}, {y_{i}}),
\end{eqnarray}
where $\mathcal{L}_{simc}$ denotes the instance discrimination loss proposed in SimCLR~\cite{chen2020simple}. We set $\lambda$ as 0.1. We gladly make the code of open-set recognition available when acceptance.

\section{Additional Results of UNDA}
We provide additional results excluded from the main paper due to limited space.

\textbf{Open-partial DA on OfficeHome.}
We show the result of open-partial DA using OfficeHome in Table~\ref{tb:officehome_opda}. \oursv outperforms baselines with a large margin in all scenarios. 

\textbf{Detailed metrics on Office.} 
In Table~\ref{tb:op_office_detail}, the accuracy of all samples (ALL), H-score, and accuracy to reject unknown samples (UNK) are described for open-set setting on Office.

\textbf{AUROC of Entropy and \ours.}
In experiments of the main paper, we show that \ours provides good thresholds to decide whether an input sample comes from known or unknown categories. In this analysis, we investigate whether the output of our open-set classifier separates known and unknown samples well. Although our focus is to build a method that determines the threshold well, this analysis will be useful to better understand our method. 
Here, we compare the entropy of classification output, the value of the predicted category's softmax output, and our one-vs-all classifier's output. To ablate the effect from unlabeled target samples, we train a model without open-set entropy minimization. 
Table~\ref{tb:auroc} shows AUROC of each metric. Although the output of \ours is not always better at calibrating uncertainty of the output than Entropy, \ours often outperforms the softmax output.

{\small
\bibliographystyle{ieee_fullname}
\bibliography{egbib}
}

\end{document}